\documentclass{article}
\usepackage{amsmath}
\usepackage{graphicx}
\usepackage{natbib}
\usepackage{multirow}
\usepackage{authblk}
\usepackage{booktabs}
\usepackage{url}

\title{Improving Short Text Classification With Augmented Data Using GPT-3}

\author[1]{Salvador V. Balkus}
\author[2]{Donghui Yan}
\affil[1]{Program in Data Science}
\affil[2]{Department of Mathematics}
\affil[ ]{University of Massachusetts Dartmouth}
\affil[ ]{\textsuperscript{1}sbalkus@umassd.edu, \textsuperscript{2}dyan@umassd.edu}
\date{}

\begin{document}
\maketitle

\begin{abstract}
GPT-3 is a large-scale natural language model developed by OpenAI that can perform many different tasks, including topic classification. Although researchers claim that it requires only a small number of in-context examples to learn a task, in practice GPT-3 requires these training examples to be either of exceptional quality or a higher quantity than easily created by hand. To address this issue, this study teaches GPT-3 to classify whether a question is related to data science by augmenting a small training set with additional examples generated by GPT-3 itself. This study compares two classifiers: the GPT-3 Classification Endpoint with augmented examples, and the GPT-3 Completion Endpoint with an optimal training set chosen using a genetic algorithm. We find that while the augmented Completion Endpoint achieves upwards of 80 percent validation accuracy, using the augmented Classification Endpoint yields more consistent accuracy on unseen examples. In this way, giving large-scale machine learning models like GPT-3 the ability to propose their own additional training examples can result in improved classification performance. 
\end{abstract}

\section{Introduction}

Text messages, social media posts, emails, and internet comments are just a few examples of ``short text'' data that people use to communicate every day \citep{Song2014}. Because of their ubiquity, building natural language processing (NLP) models that can classify the topic or category of short text is an important business problem \citep{Li2020}. For example, classification models could automatically detect and censor offensive text, identify when customers ask questions that should be forwarded to a specific contact, or determine the sentiment of a message - whether it is considered positive or negative. Previously, building text classification models was challenging as it required collecting vast quantities of training examples to learn subtle differences in context - a quantity that may not have been feasible to collect for individuals, academics, or small businesses. 

In recent years, however, new developments have made text classification accessible for everyone. \textit{Transfer learning} is a process by which researchers pre-train a large neural network to be easily tuned or generalized to many different types of problems. By training models on vast quantities of unstructured text data from the internet, artificial intelligence research companies have developed transfer learning models for natural language processing problems. One such model is the Generative Pre-trained Transformer 3 (GPT-3) which was recently made commercially available to the public by OpenAI \citep{Brown2020, Dale2020, Pilipiszyn2021}. 

Because their ability to generate human-sounding text can be generalized to many different tasks, GPT-3 and other transfer learning models are poised to solve numerous natural language problems. In addition, GPT-3 is described as ``few-shot learning'' meaning that only a few training examples are required to teach the model to perform a task. As a result, GPT-3 allows developers with limited data to develop applications that rely on text classification models. Developers have already created dozens of applications using GPT-3, and this number will likely only grow \citep{Pilipiszyn2021}. 

However, despite how it has been advertised, our research indicates that the few-shot learning ability of GPT-3 is limited for text classification. There are two methods by which the GPT-3 API provides classification capabilities. One is the \textbf{Completion Endpoint}, which requires a text prompt followed by example-label pairs as input. In this case, the model only requires a few (5-10) examples, but its performance is highly sensitive to which examples are included in the context. The other is the \textbf{Classification Endpoint}, which requires a JSON file of example-label pairs. This endpoint is more reliable, but requires dozens, hundreds, or even thousands of examples to perform well - far more than would typically be considered ``a few.'' Hence, those who do not have access to large text datasets for their specific problem require a way to use GPT-3 and other transfer learning models for classification when data quality or quantity is limited. 

This raises the research question of this study: how can we improve training examples for GPT-3? This process is often called \textit{data augmentation}, and we are especially interested in its practical application to empirical text classification problems faced by organizations with small datasets.

This study evaluates two potential data augmentation methods. The first involves using GPT-3 to generate new questions of a given class based on the existing ones. This increases the \textit{quantity} of the sample size for the Classification Endpoint, with the goal of improving accuracy. The second optimizes which training examples are included in the Completion Endpoint using a genetic algorithm, in hopes of improving the \textit{quality} of the input context.

We evaluate these two methods in a case study: predicting question topics related to data science. At the University of Massachusetts Dartmouth, a student organization called the Big Data Club collected questions that students asked through the organization's server on Discord, an instant messaging app. To help track which members participated the most in the group, Big Data Club wanted to use GPT-3 to identify when members asked questions related to data science, the group's area of interest. Using data from this case study, our research evaluates the performance of two data augmentation methods with the GPT-3 Classification and Completion Endpoints for solving a practical, real-world, text classification problem. 

By providing methods to train natural language models on small short text datasets, our research benefits developers looking to apply GPT-3 and other pre-trained generative transfer learning models in practice. Although some text data augmentation techniques have already been discussed in the literature, they are limited in scope, do not use GPT-3, and are not evaluated on classification beyond common NLP benchmarks. Overall, the goal of this study is to evaluate whether GPT-3 can be used to improve its own performance, rather than relying on external methods. This would promote the implementation and more widespread use of natural language models for automating mundane business tasks requiring text classification, from commercial chatbots to the detection of harmful or misleading social media posts. 

The remaining content is summarized as follows. Section \ref{lit} reviews the existing literature on data augmentation for NLP problems and provides readers with necessary background information on GPT-3 and the other methods we use for augmentation. In Section \ref{methods}, we explain the workings of the proposed augmentation methods implemented in this study. Section \ref{eval} describes how these methods were evaluated, including the data and model parameters. Then, Section \ref{results} presents our evaluation results, while a discussion on these results, their limitations, and potential future work is included in Section \ref{discuss}. Finally, Section \ref{conclusion} summarizes our conclusions.

\section{Background and Related Work}
\label{lit}

\subsection{Natural Language Processing and Transfer Learning}

Text classification is defined as ``the procedure of designating pre-defined labels for text'' \citep{Li2020}. Language classifiers take tokens - the most basic components of words - as input in order to yield labels as output. Previously, traditional machine learning models such as the Naive Bayes, Support Vector Machine, K-Nearest Neighbor, and Random Forest algorithms have been combined with text representation methods such as N-gram, Bag-Of-Words, and word2vec for text classification \citep{Kowsari2019, Li2020}. However, these methods often require feature engineering and suffer performance limitations. Classification is especially challenging for short text because of the sparseness of the training examples, which contain very few words to inform the model \citep{Song2014}.

More recent research focuses on deep learning, which uses large-scale neural networks to better capture the complex, nonlinear relationships between words \citep{Zulqarnain2020, Minaee2022}. However, deep neural networks often require time-consuming and expensive training processes to develop. To overcome this problem, researchers have implemented the technique of transfer learning. Transfer learning involves constructing a pre-trained deep learning model for some general task, which can then be fine-tuned for a specific task using a smaller dataset \citep{Dai2015, Radford2018}. Using pre-training reduces the time and data necessary to achieve quality performance in a model. It can even allow the construction of ``few-shot'' learning models which require only a small number of training examples to attain acceptable performance. 

Transfer learning models for NLP are trained on one task: predicting the next token in a sequence of text. This can be thought of as analogous to the human process of writing, as it allows the model to generate text by repeatedly predicting a sequence of tokens given an input. To develop a transfer learning model, a neural network with millions or billions of neurons is trained on a large corpus of unstructured text data, usually from the internet, to predict masked tokens on unseen passages. This allows the model to, in essence, learn the relationships between thousands of different words in different contexts. Then this model, in turn, can be applied to numerous natural language problems either through fine-tuning or by providing a prompt structured in a specific pattern, which the model will attempt to continue \citep{Devlin2019, Brown2020}.

Some transfer learning models for NLP include BERT \citep{Devlin2019}, which has spawned numerous spin-off models \citep{Liu2019, Adhikari2019}; GPT-3 \citep{Brown2020}, which has been deployed for dozens of commercial products; and even more cutting-edge models like ERNiE \citep{Sun2021}, ST-MoE \citep{Zoph2022}, and Turing NLR \citep{Bajaj2022}. Our paper focuses on GPT-3, since it is currently commercially available in public beta through an API, allowing anyone to use it for practical applications.

\subsection{GPT-3}
GPT-3 is a recent transfer learning model developed for natural language problems \citep{Brown2020}. Its deep neural network architecture features layers of \textit{transformers}, which are deep learning layers that use self-attention (modeling the relationship between each word and all other words in the sequence) to learn the complex relationships between words in different contexts \citep{Vaswani2017}. Several versions of the model were trained, with the largest having up to 175 billion neurons. The model was trained on the Common Crawl dataset \citep{Brown2020, Raffel2019} which contains unstructured text data scraped from the web, including sites like Wikipedia.

GPT-3 is commercially available through an application programming interface (API) offered by OpenAI \citep{Pilipiszyn2021}. It features several different engines, of which this study considers two: \textbf{ada} and \textbf{davinci}. The \textbf{ada} engine provides the fastest and cheapest inference but limited predictive performance, while \textbf{davinci} offers the most accurate predictions, but is the most expensive and is limited by slow inference time. 

Short text classification can be performed using the OpenAI GPT-3 API in two main ways. The first is the \textbf{Completion Endpoint}, which is traditionally used for generating human-sounding text given a prompt \citep{Completion}. However, by structuring the prompt in a specific pattern, the GPT-3 Completion Endpoint can also be used for classification. If users provide a prompt that consists of a series of examples, each followed by a label, and leave the label of the final example blank the Completion Endpoint will follow the pattern provided by the user and attempt to predict the label of the final example. Users can also restrict the output to ensure the only possible token outputs are the known classes. This process is explained further in Section \ref{comp-aug} and depicted visually in Figure \ref{met:interface}. However, because the input is restricted to 2,048 tokens and more tokens induce greater expense, the number of possible training examples that can be used with the Completion Endpoint is limited. 

The second short text classification method with GPT-3 is the \textbf{Classification Endpoint}. This method, designed specifically for classification, requires uploading a JSON file containing training examples to the OpenAI API. When called to classify a short text input, the Classification Endpoint searches for examples semantically similar to the input, ranks them based on their relevance, and then selects the class with the highest likelihood of occurring based on the labels of the most similar examples \citep{Classification}.

Both the Completion and Classification Endpoints enable users to develop \textit{few-shot} natural language classification models - in other words, GPT-3 requires only a ``few'' examples to learn a task, rather than the terabytes of data necessary for the original pre-training. At the time of release, GPT-3 had achieved state-of-the-art performance on the SuperGLUE natural language benchmark tasks \citep{Brown2020}, including reading comprehension and question answering. Current research has applied GPT-3 to other tasks, such as understanding and drafting emails \citep{Thiergart2021} It also attracted media attention due to its human-like performance in writing fake news articles \citep{Guardian2020}. Because of the potential ethical issues raised by the model, OpenAI closely restricts its deployment to only approved users and use cases \citep{Pilipiszyn2021}.

Despite its promise, using GPT-3 is still challenging. The performance of GPT-3 highly depends on the choice of examples provided in the few-shot learning scenario \citep{Liu2021, Zhao2021}. Hence, a method for training example selection is necessary. \textit{Contextual calibration} provides a method to select where in the prompt to place different answers to avoid instability in responses \citep{Zhao2021}. However, this method only selects \textit{where} to place the example, not which example to use. The proposed method KATE selects optimal in-context examples for the Completion Endpoint by retrieving examples semantically similar to the test sample before constructing the prompt for GPT-3 \citep{Liu2021}. However, this requires modifying the prompt of GPT-3 for each prediction, may result in very expensive API calls, and is already performed by the Classification Endpoint.

As such, a better method for improving the performance of GPT-3 when many high-quality examples are not available is necessary. For high-quality examples to be selected, they must exist in the first place. What happens if the user does not have any high-quality examples? In this case, data augmentation methods can be used to improve the performance of GPT-3. 

\subsection{Data Augmentation}
Data augmentation is the process of generating additional observations to increase the size of the training data for a machine learning model. Previous research has proposed several techniques, though compared to other domains such as image processing \citep{Shorten2019}, augmentation of text data is still in its infancy.

Firstly, some text data augmentation techniques focus on modifying existing text to increase the number of samples. For instance, in the SentAugment technique, previously unlabeled sentences are retrieved from a text bank and labeled to increase the number of training examples in the model \citep{Du2020}. Additionally, \cite{Qu2020} discuss several variations of text data augmentation which rely on adversarial training methods to generate new examples. They also propose CoDA, which combines data transformations to augment training data. Though successful, these methods are complex to implement and do not necessarily allow the model to improve by proposing examples that are \textit{better} than the existing ones. 

Alternatively, other methods focus more on creating brand-new text altogether. Such methods rely heavily on the generative capabilities of transfer learning models to create new examples. For example, GPT3Mix uses GPT-3 to select samples from training data and generate additional samples by mixing previous sentence components together into new, yet still plausible, examples \citep{Yoo2021}. More broadly, \cite{Kumar2020} use a variety of NLP transfer learning models to generate new examples from existing ones for augmentation. The approach improved performance on sentiment analysis, intent classification, and question type classification. However, this approach relied on NLP benchmarks rather than testing on a practical problem and was not tested using GPT-3. 

Therefore, inspired by \cite{Kumar2020}, one major contribution of this research is to test whether generating new examples using transfer learning is effective on practical problems beyond those contained in NLP benchmark datasets. While we also perform data augmentation using the generative capabilities of transfer learning, our research extends these methods to be applicable to the OpenAI GPT-3 API Endpoints for classification. This allows leveraging of the newer GPT-3 model which has displayed much better text generation capabilities. We also explore the performance of data augmentation in a more practical, empirical case study with highly limited data availability. Furthermore, we propose and examine optimization methods that select the best examples for the GPT-3 Completion Endpoint, whose input prompt is limited to only a few examples. This process is performed using a genetic algorithm. 

\subsection{Genetic Algorithms}

As discussed earlier, GPT-3 is sensitive to the in-context examples selected as training data - especially the Completion Endpoint, which uses limited examples since larger inputs are more expensive and their size is capped at 2,048 tokens. Using the GPT-3 Completion Endpoint for few-shot natural language classification requires selecting the best examples to include in its context. This amounts to an optimization problem wherein the feature space is defined by text examples rather than numeric values. Because of this, traditional optimization algorithms such as gradient descent cannot be applied. Instead, we select optimal examples using a genetic algorithm.

A \textit{genetic algorithm} is an iterative optimization technique inspired by the biological mechanism of natural selection. The algorithm initializes and maintains a population of potential solution candidates. At each iteration (termed \textit{generation}) the algorithm evaluates the \textit{fitness} of each candidate, which is the value to be optimized. The candidates with the highest fitness are preserved and their genetic information is recombined using a crossover operator to produce genetically similar offspring, which are then evaluated at the next iteration. As this process continues, the fitness of the population rises, thereby maximizing the fitness function \citep{Srinivas1994, Katoch2020}. 

In a genetic algorithm, each candidate is defined as a set of \textit{alleles}, each of which describes a feature of the candidate. Genetic algorithms also feature several \textit{operators} applied at each iteration, which are applied in the following order \citep{Srinivas1994, Katoch2020}:
\begin{enumerate}
    \item \textit{Encoding}. Candidate solutions can be encoded in multiple ways, such as binary digits or, in the case of this study, a value like a string of characters. 
    \item \textit{Fitness Evaluation}. At the start, each possible candidate is evaluated based on the function to optimize. For example, when seeking to optimize a machine learning algorithm, a performance metric like accuracy or even F1 score can be used as the fitness function.
    \item \textit{Selection}. This operator selects the best candidates to provide offspring based on their fitness. There are many selection mechanisms, including rank-based as well as \textit{tournament selection}, the method used in this study. In tournament selection, groups of candidates are randomly created, and the candidate with the best fitness survives. In addition, \textit{elitist} selection allows previous candidates to remain in the population.
    \item \textit{Crossover}. The process by which remaining candidates produce offspring by combining their sets of alleles into one new set. \textit{Partially-matched crossover} is the most common for sequence data (like text), as it preserves the order of observations. 
    \item \textit{Mutation}. To avoid premature convergence, some alleles are randomly modified using a technique appropriate to the data representation. For example, a new value can be randomly sampled from existing possible values.
\end{enumerate}

Genetic algorithms have been applied to numerous problems, such as logistics, information security, image and video processing, agriculture, gaming, and wireless communications \citep{Katoch2020}. Genetic algorithms are useful in NLP because they can perform optimization on functions that do not take numbers as input - including those which take text, as in the problem faced by this study. This allows them to optimize natural language training examples for few-shot NLP learning models like GPT-3. We use a genetic algorithm to select the optimal examples for the GPT-3 Completion Endpoint.

\section{Proposed Augmentation Methods}
\label{methods}

This study evaluates two methods for improving short text classification by augmenting training data using GPT-3. The first augments the GPT-3 Classification Endpoint, which classifies each short text input by searching for the most semantically similar training examples and comparing the probabilities of each class. In this method, additional training examples for the Classification Endpoint are generated using GPT-3 and added to the set of training examples in an attempt to improve performance. The second method augments the GPT-3 Completion Endpoint using a genetic algorithm to select the optimal examples to be included in the input context. These methods are described in detail as follows.

\subsection{Classification Endpoint Augmentation}
\label{class-aug}

As mentioned in Section \ref{lit}, the OpenAI GPT-3 Classification Endpoint performs classification by comparing the input text to a labeled training set. After a semantic search to identify $n$ examples, the most semantically similar results are ranked based on their relevance. Then, a label is generated based on the probabilities of the labels for the selected results. This process is provided by the OpenAI API \citep{Classification}.

\begin{figure}[!h]
\includegraphics[width=4.5in]{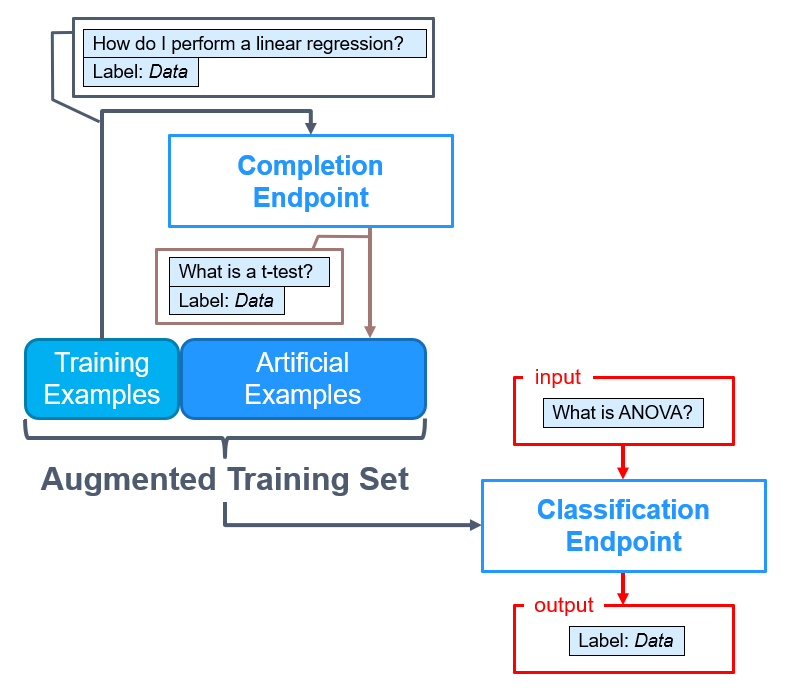}
\centering
\caption{Our Classification Endpoint Augmentation. First, new artificial questions are created using the ability of the GPT-3 Completion Endpoint to generate text based on existing examples. Then, newly generated questions are used to train the Classification Endpoint to, given an input, produce a more accurate output. }
\label{met:classification}
\end{figure}

To augment the GPT-3 Classification Endpoint, this study simply generates additional training examples using GPT-3 itself - specifically, using the GPT-3 Completion Endpoint, which is designed to generate text given a prompt. To generate each additional training example for Classification Endpoint Augmentation, we first provide the GPT-3 Completion Endpoint with the prompt ``Generate a similar question:'' followed by three questions of the same class (``data'' or ``other'') each preceded by the ``Q:'' token. These questions are randomly selected from the original training set. Finally, the output question is labeled with the same class as the input questions. The process is repeated to generate the desired number of questions with which to augment the Classification Endpoint. This is similar to the method used in \citep{Kumar2020}; the difference is that here GPT-3 is used to ``teach itself'' by providing its own generated input examples. Figure \ref{met:classification} demonstrates this process graphically.

Once the desired number of example questions are generated, they are added to the training set, which is then uploaded to the OpenAI Classification Endpoint. With this training set, represented as a JSON file, users can call the GPT-3 API to classify any given text input. Of course, to enhance performance, users should compare the performances of models that use different numbers of additional training examples, as well as test different parameters for the Classification Endpoint such as \textit{temperature} and \textit{max\_examples} - these are known as \textbf{hyperparameters}. There are many ways to optimize hyperparameters \citep{Bischl2021}; the methods used in this study are described in Section \ref{eval}.

\subsection{Completion Endpoint Augmentation}
\label{comp-aug}

\subsubsection{Classification using the Completion Endpoint}

\begin{figure}[!h]
\centering
\includegraphics[width=4.5in]{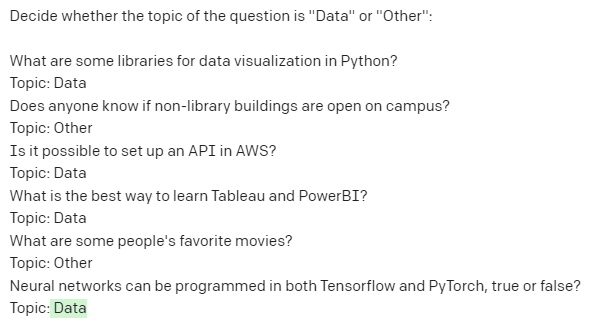}
\caption{Example of an input to the GPT-3 Completion Endpoint interface. By organizing text in the \textit{question-topic-question-topic} pattern, GPT-3 can be instructed to output labels classifying the topic of a question. The output of the model is highlighted.}
\label{met:interface}
\end{figure}

In the first augmentation method we just described, we use the Completion Endpoint to generate new examples to attempt to augment the Classification Endpoint. Recall from Section \ref{lit}, however, that the Completion Endpoint can also be directly applied to classification problems. To classify the topic of a question, we prompt the Completion Endpoint with the phrase ``Decide whether the topic of the question is `Data' or `Other''', and append several training examples in a \textit{question-topic-question-topic} pattern. An example of this input to the Completion Endpoint that results in classification is shown in Figure \ref{met:interface}. 

The examples provided, often referred to as ``in-context examples,'' serve as a miniature training set for the Completion Endpoint. To classify a question, we simply append it to the end of the prompt and add a ``Topic:'' token with no label after it. When provided as input, this prompts GPT-3 to predict the next token as the topic of the previous question. The API even allows us to restrict the output of GPT-3 to specified tokens, so we can ensure that only the possible classes (in this case, ``Data'' or ``Other'') can be output as predictions. Hence, the Completion Endpoint can be used as a classifier after being provided with only a few training examples. 

While the Completion Endpoint can also be provided augmented training examples in the same way as the Classification Endpoint, input prompts are limited to a certain number of tokens, which means only a small number of training examples can be used. In addition, limited training examples are desired because the more tokens included in the prompt, the more expensive the model is to call from the API. Therefore, rather than augmenting the Completion Endpoint by generating additional training examples, this study uses an optimization algorithm to select which subset of examples from a larger training set yields the best accuracy when provided as a prompt.

\subsubsection{Performing Augmentation By Optimizing Training Example Selection}

\begin{figure}[!h]
\includegraphics[width=4in]{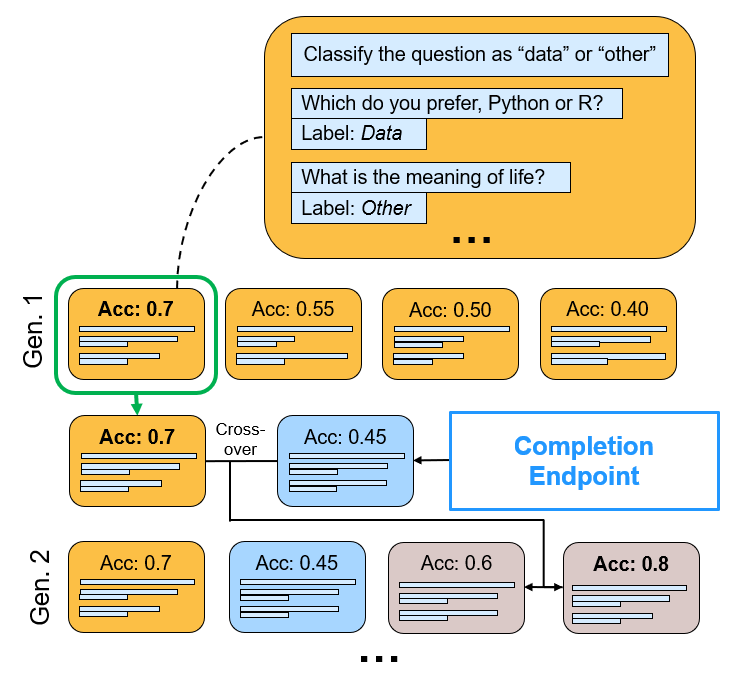}
\centering
\caption{Graphical depiction of the genetic algorithm for selecting optimal augmented in-context examples for the GPT-3 Completion Endpoint. Each candidate consists of a set of alleles representing questions provided to the Completion Endpoint prompt. At each generation, the candidates with the best accuracy are selected to produce offspring with new candidates containing augmented examples generated by GPT-3. }
\label{met:genetic-diagram}
\end{figure}

To optimize which augmented examples are chosen to include in the Completion Endpoint prompt, this work employs a genetic algorithm, as described in the Section \ref{lit} literature review. Recall that a genetic algorithm maintains a population of candidates with certain alleles (traits). Here, each candidate represents a possible prompt to provide to the Completion Endpoint, and each allele represents a single example-label pair. Hence, alleles are encoded as strings of text. The gene pool, then, is the set of all possible training example-label pairs. The genetic algorithm process is depicted in Figure \ref{met:genetic-diagram}.

\begin{figure}[!h]
\includegraphics[width=4in]{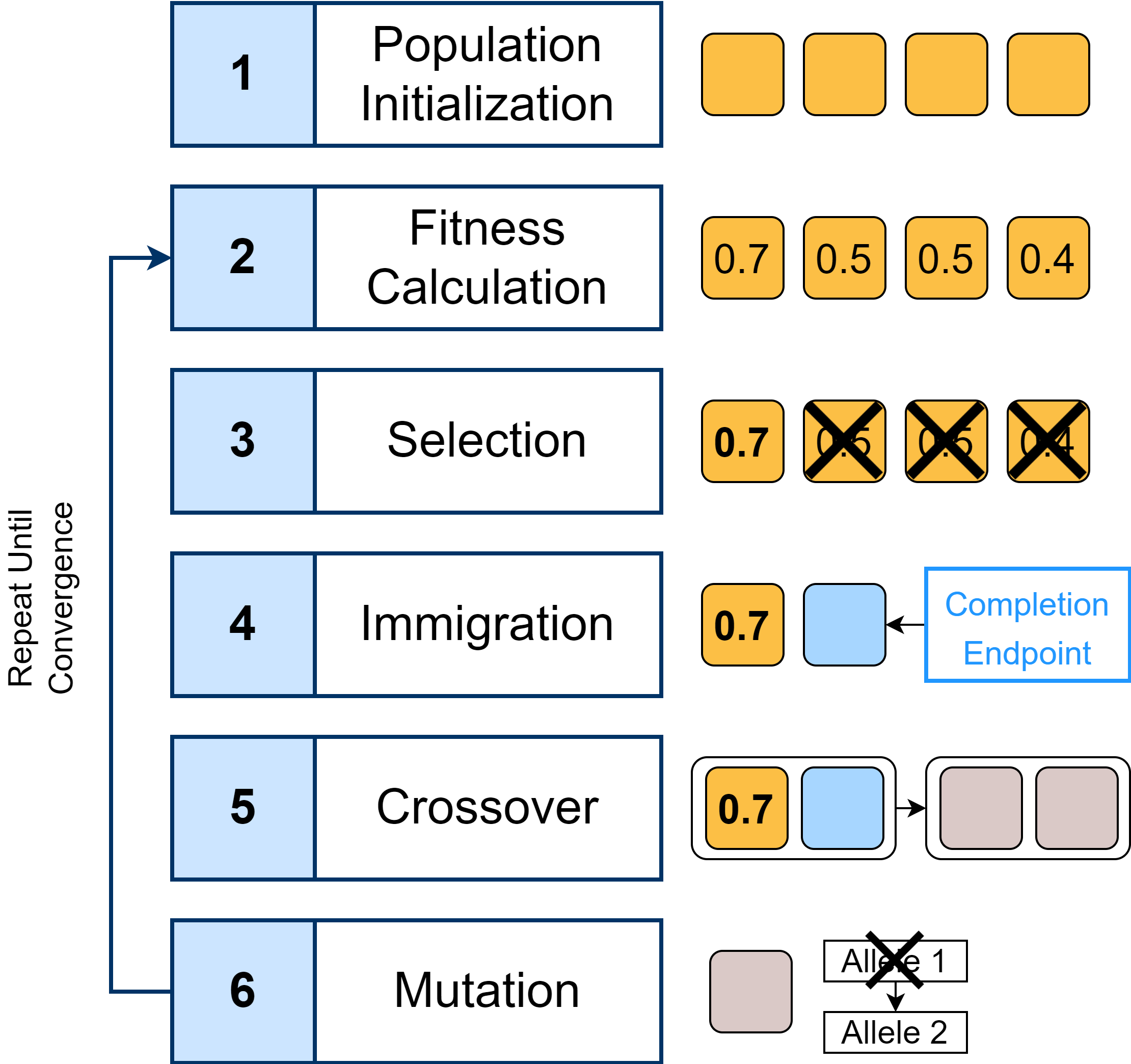}
\centering
\caption{Genetic algorithm steps for Completion Endpoint Augmentation.}
\label{met:steps}
\end{figure}

As mentioned previously, a genetic algorithm applies a number of operators at each iteration. Operators apply some transformation to the existing population of candidates, and at each iteration, the fitness of the candidates improves. The specific operators we use in the genetic algorithm for optimizing Completion Endpoint training examples are described as follows (note that Step 1 is applied only in the first iteration), and are listed in Figure \ref{met:steps}.

\textit{Step 1: Population Initialization.} To begin, each candidate in the population is initialized as a set of random alleles sampled from the training set. We sample the same number of alleles from each class (``Data'' or ``Other''). The alleles are sampled uniformly without replacement (only being replaced when the training set is empty), as the algorithm should ensure that no duplicate alleles are placed together in the same candidate (which would be inefficient). Then, each candidate is evaluated using the fitness function.

\textit{Step 2: Fitness Calculation.} The fitness function is defined as GPT-3's predictive performance on the validation set. In this study, accuracy is used, but other measures such as sensitivity, F1 score, \textit{et cetera} are also possible. To evaluate the fitness, each candidate is used to predict the label of each observation in the validation set, and its validation accuracy is calculated as the fitness. Hence, we seek to maximize candidate accuracy.

\textit{Step 3: Selection.} Once a fitness has been calculated for each candidate, selection is performed using \textit{4-way tournament selection}. Candidates are allocated randomly into groups of four, and only the candidate with the best accuracy moves on to the next generation. This ensures that the algorithm uses elitist selection, allowing the best individual from the previous population to carry over to the next population. This prevents the best solutions from being lost. However, its random nature also promotes genetic diversity by allowing a small number of candidates that are not the best to possibly survive as well. 

\textit{Step 4: Immigration.} For such a small training set, it is necessary to introduce augmented examples - otherwise, the lack of genetic diversity will cause the algorithm to converge prematurely. This is performed using \textit{immigration}, an operator which introduces candidates with entirely new sets of alleles into the gene pool \citep{Yang2004}. The immigration operator in this algorithm employs the GPT-3 Completion Endpoint to generate new alleles in the same manner as in Classification Endpoint data augmentation, as discussed in the previous section. 

\textit{Step 5: Crossover.} Genetic algorithms optimize by proposing new candidates likely to have high accuracy at each iteration. This is done by creating ``offspring'' which combine observations from different existing candidates. In this study, each candidate selected in Step 3 performs crossover with an immigrant to generate two offspring candidates. We accomplish this using partially-matched crossover which randomly selects one or more alleles to swap between the sets of alleles of the selected candidate and the immigrant \citep{Katoch2020}. These alleles are selected in such a manner as to prevent duplication of the same allele. The newly created sets become the alleles for each offspring (the process is performed twice, once for each). 

\textit{Step 6: Mutation.} Finally, each allele in the offspring has a chance to be modified with some mutation probability. The mutation operator replaces the allele with a random allele sampled from the total gene pool - the set of all alleles in the population, including those newly generated in each immigrant population and excluding those already contained in the candidate. This promotes genetic diversity and prevents the optimization algorithm from being limited to the local optima of the best candidates found so far. 

After this process, a new population is created which, provided the original was divisible by 4, contains the same number of candidates as the original. Then, steps 2 through 6 are repeated for the desired number of iterations (termed ``generations'') until the best candidate's desired accuracy is achieved, or some other stopping criteria such as a time or spending limit is met. 

By performing this procedure, the algorithm causes the best candidates to reproduce with new candidates, searching for new possible solutions that are similar yet different to the best found so far. Using mutation and immigration to preserve genetic diversity prevents the population from becoming overrun with candidates that have the same alleles, thereby avoiding convergence to a mere local optimum. By optimizing the prompt for accuracy, we select the best examples for the Classification Endpoint - the ones which yield the best performance.

\section{Numerical Evaluation}
\label{eval}

This section describes how the above methods were evaluated to compare their performance, including data collection and implementation. Though the overall algorithms for data augmentation were described previously, here we detail which specific parameters were used for the models and the basic reasoning behind their selection in the evaluation process. 

\subsection{Data Collection}

To train the model, we collected a dataset of short text from the University of Massachusetts Dartmouth. This dataset consisted of questions asked in the Discord instant messaging app by undergraduate and graduate members of the University of Massachusetts Dartmouth Big Data Club. The Big Data Club Discord server is used by students at the university to specifically discuss data science topics as well as engage in casual conversation over text-based chat. As a result, we were able to collect both questions related to data science as well as questions related to other topics.

The dataset contained 72 questions that we labeled either ``data'' or ``other'' to indicate whether the topic was related to data science. Of these, 45 were collected directly from messages in the Discord server, with the members' permission. The remaining 27 were proposed by club members aware of the research project and edited by the research team with the goal of covering a broad range of topics in data science, such as statistics, machine learning, databases, and cloud computing. They also included counterintuitive examples that use data science terms in a non-data-related context, as well as ``junk'' questions. For example, a counterintuitive question might be ``How many neurons are contained within the human nervous system?'' - although neural networks are a type of machine learning model, in this context the term ``neuron'' does not relate to data science. An example of a junk question would be a single-word question like ``What?'' or a string of random characters. Not only did these expand the distribution of possible topics represented in the training set, but they may also have served as ``adversarial examples'' which would have been exceptionally hard to classify or that would have represented boundary conditions in the data.

These questions were divided into three sets. The \textbf{training set} and \textbf{validation set} contained 26 questions each, with a random allocation of questions that occurred naturally and that were proposed by club members. The \textbf{test set} contained 20 questions and included only questions that were originally asked in the Discord server. All sets were selected randomly to contain the same number ``data'' and ``other'' questions to avoid class imbalance issues.

The training set was used as input to each endpoint; it included the examples from which the algorithm actually learned. The validation set was used to evaluate performance and optimize parameters, such as the examples selected for the prompt in the genetic algorithm. The test set was also used to evaluate performance - it represented questions that the algorithm had never seen before, so performance on this set is exceptionally notable.

\subsection{Implementation}

Data processing, experimentation, and programming of the genetic algorithm were performed in Python 3.8.5. The Python API for OpenAI was used to call predictions from GPT-3. Except for example generation, all GPT-3 models used the \textbf{ada} engine since it is currently the least expensive and yielded the fastest inference time - necessary characteristics for a real-time deployed machine learning application. New questions were generated using the \textbf{davinci} engine since it is optimized by performance. After data augmentation, once enough questions were generated, the model could be used continuously without needing to generate any more questions, negating any considerations of speed or cost. This is why we used davinci for generating new questions during data augmentation, but not for the actual classification task.

\subsection{Model Parameters}

\subsubsection{Classification Endpoint Augmentation}

To evaluate augmentation on the Classification Endpoint Augmentation, we ran a battery of tests. In each test, $n$ question-label pairs were first sampled from a set of 11,000 questions generated by the GPT-3 Completion Endpoint (using the procedure specified in Section \ref{class-aug}) and added to the 26-question training set to create the augmented training set. Then, the augmented training set was formatted into a JSON file and uploaded to the API. Finally, using each augmented training set, we evaluated the performance of the model on the validation set and the test set. Tests were run for $n=$ 0 (for a baseline), 10, 100, 1,000, and 10,000 augmented examples added. 

Each test was repeated 5 times. All augmented Classification Endpoint models were first evaluated on the validation set using different sets of hyperparameters to determine which hyperparameters were the best. We optimized two hyperparameters, \textit{temperature} and \textit{max\_examples}, using a grid search \citep{Bischl2021}. For \textit{temperature}, we tested values of 0, 0.1, and 0.5, and for \textit{max\_examples}, we tested values of 5, 10, 15, 20, 25, and for $>$100 added examples, 100. 

After this, we selected the set of hyperparameters that achieved the best average accuracy on the validation set. A model using these hyperparameters was then evaluated 5 times on the test set. This provides a description of the model's performance on unseen data, which more closely approximates how well it would perform if it were deployed in practice. 

\subsubsection{Completion Endpoint Augmentation}

To evaluate the augmentation method for the Completion Endpoint, we employed the genetic algorithm described in Section \ref{comp-aug} to optimize the endpoint's accuracy on the validation set. The genetic algorithm was run for 40 generations, and this test was repeated 3 times. We chose these quantities to balance assurance of reproducibility with available funds, which limited the total number of generations that could be run. 

Table \ref{gen_params} displays the parameters used in the genetic algorithm. The fitness function calculated the accuracy of the algorithm on the validation set. The model's accuracy on the validation set is chosen as the fitness since it is easily interpretable. Accuracy is also commonly used for NLP model performance evaluation \citep{Brown2020}. Temperature was set to 0 to ensure results are deterministic. Partially-matched crossover ensured that no duplicate examples were placed into the context \citep{Katoch2020}. The mutation rate of 0.1, although high, was chosen to ensure that most offspring would obtain at least 1 mutation since there are only 8 alleles to be mutated. 

\begin{table}[!h]
\begin{center}
    \caption{Genetic algorithm parameters for Completion Endpoint in-context example selection optimization.}
    \label{gen_params}
    \vspace{2mm}
    \begin{tabular}{cccc}
    \toprule
    Parameter & Value & Parameter & Value \\
    \midrule
    Encoding & String & Population size &  32  \\
    Selection method & Tournament & Tournament size & 4 \\
    Crossover method & Partially-matched & Crossover probability & 1.0 \\
    Mutation method & Uniform & Mutation rate & 0.1 \\
    Fitness function & Accuracy & Validation set size & 26  \\
    GPT-3 Engine & Ada & GPT-3 Temperature & 0 \\
    Header & Yes & Number of alleles & 8 \\
    \bottomrule
    \end{tabular}
\end{center}
\end{table}

Regarding the population dynamics, in each of the three tests, a population of 32 candidates with 8 alleles (question-label pairs) each was randomly initialized by sampling sets of alleles from the 26 training questions, with no duplicate questions permitted. These numbers were chosen to ensure that different combinations of the sample examples could be represented in the starting population. This also ensured that when each tournament winner produced 2 offspring with a new immigrant, the original population size would be recovered. At each generation, after 4-way tournament selection culled all but 8 candidates, 8 new immigrants were introduced, and crossover yielded 16 new offspring; as a result, we attain the same population size (32) as that prior to selection.  

After plotting the model's accuracy on the validation set over time, we selected the best candidate in terms of fitness (accuracy) that occurred at the end of the 40 generations. We then took the examples from the best candidate as the Completion Endpoint model. Finally, the accuracy of this model was evaluated on the test set. This yielded the algorithm's performance on examples it had never seen before, to ensure that it continued to perform well in practice. The performance of our augmentation methods on the validation and test sets is described in the next section.

\section{Results}
\label{results}

\subsection{Classification Endpoint Augmentation Performance}

Data augmentation for the Classification Endpoint successfully resulted in increased model accuracy on text classification for the problem of classifying whether a question is related to data science. Table \ref{res:tab-class} reports the mean $\mu$ and standard error of the endpoint accuracy for both the validation set and the test set, with the model using optimized hyperparameters. The $p$-value reports $t$-test results comparing the mean of the baseline (0 examples added) to each scenario with $n$ examples added, to examine whether each difference in accuracy is statistically significant. 

As more generated examples were added to the training set, the Classification Endpoint accuracy tended to increase. Without augmentation, the Classification Endpoint with just the 26-question training set performed comparably to random guessing, only classifying about 49 percent of questions correctly on average for the validation set and 58 percent on average for the test set. However, on the validation set, accuracy continually increased as more examples were added, reaching about 73 percent accuracy after adding 10,000 new examples.

\begin{figure}[!htb]
\includegraphics[width=4.4in]{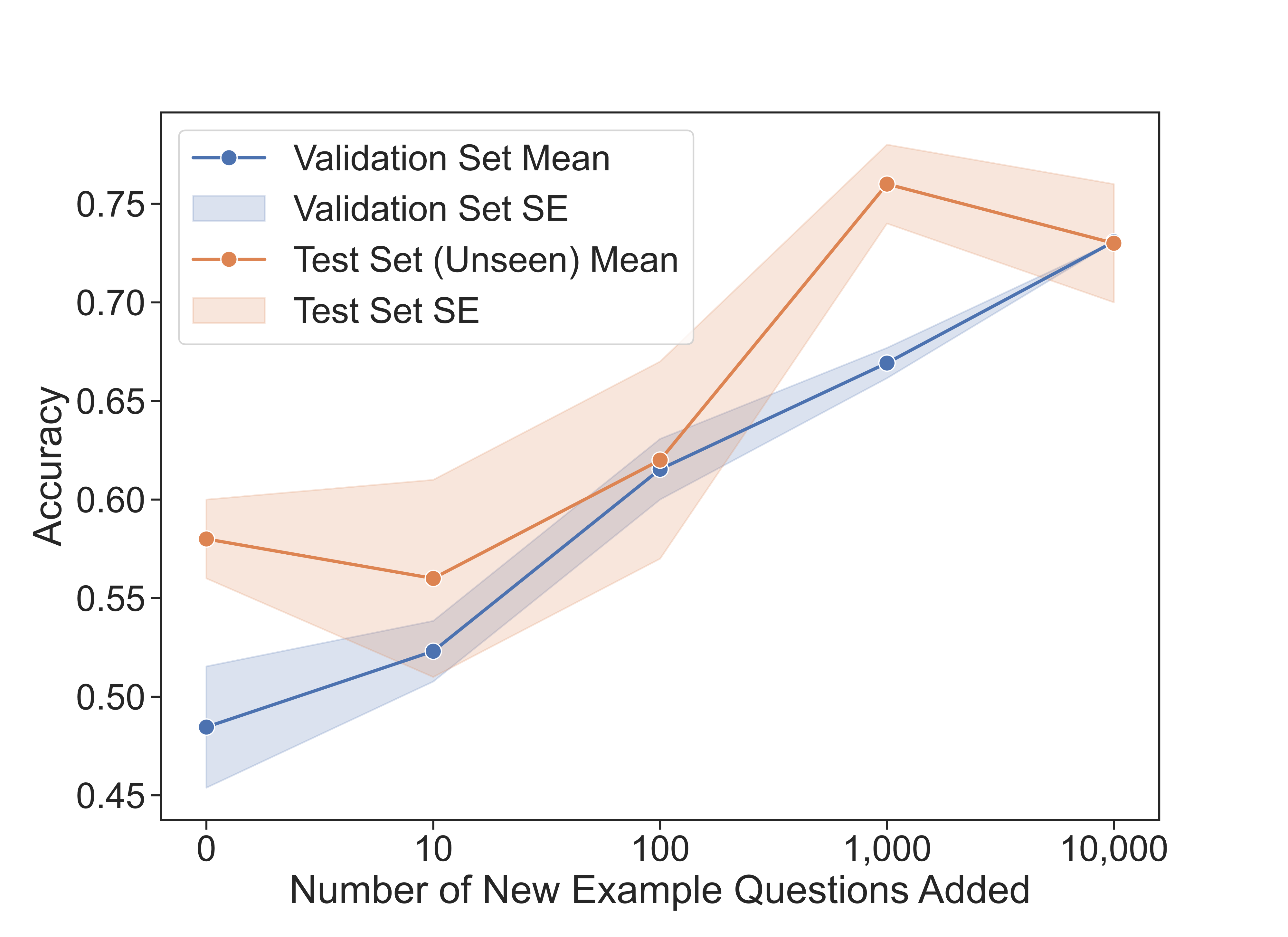}
\centering
\caption{GPT-3 Classification Endpoint mean performance with standard errors on data science question topic classification. Training data is augmented by adding different quantities of new examples generated with GPT-3 Davinci Completion.}
\label{res:fig-class}
\end{figure}

While, for the validation set, accuracy was positively related to the number of questions generated, the same was not true for the test set. Figure \ref{res:fig-class} plots the relationship between accuracy and number of example questions added across both validation and test sets, with the shaded regions representing the standard error (68 percent confidence interval). Note that the x-axis is a log scale. On the test set, accuracy scarcely increased at all until the number of questions added reached about 1,000, at which point it increased to 76 percent. This represented peak accuracy; augmented training sets with 10,000 new questions averaged only 73 percent accuracy, a slight drop.

Overall, data augmentation yielded statistically significant increases in accuracy based on $t$-tests comparing each set of $n$ additional artificial examples to the baseline. On the validation set, only 100 examples were necessary to observe statistically significant increases in accuracy at $\alpha = 0.05$. However, adding 100 examples did not yield significant increases in accuracy on the test set, as the baseline accuracy was much higher. With the more strict $\alpha = 0.01$ requirement for significance, the increase in accuracy became statistically significant after adding 1,000 examples or more for both the validation and test set.

\begin{table}[!htb]
    \caption{GPT-3 Classification Endpoint performance on data science question topic classification, additional examples generated using GPT-3 Davinci Completion. $p$-values test for significance from results with 0 additional examples.}
    \label{res:tab-class}
    \vspace{2mm}
    \centering
    \begin{tabular}{ccccccc}
        \toprule
         \multirow{2}{*}{$n$ Added Examples} & \multicolumn{3}{c}{Validation Accuracy} & \multicolumn{3}{c}{Test Accuracy} \\
         & $\mu$ & SE & $p$ & $\mu$ & SE & $p$ \\
         \midrule
         0 & 0.49 & ($\pm$ 0.028) & - & 0.58 & ($\pm$ 0.023) & - \\
         10 & 0.52 & ($\pm$ 0.018) & 0.328 & 0.56 & ($\pm$ 0.048) & 0.744 \\
         100 & 0.62 & ($\pm$ 0.022) & 0.012 & 0.62 & ($\pm$ 0.041) & 0.471 \\
         1000 & 0.67 & ($\pm$ 0.008) &  0.001 & 0.76 & ($\pm$ 0.017) & 0.001 \\
         10000 & 0.73 & ($\pm$ 0.000) & 0.001  & 0.73 & ($\pm$ 0.030) & 0.008 \\
         \bottomrule
    \end{tabular}
\end{table}

\subsection{Completion Endpoint Augmentation Results}
\begin{figure}[!h]
\includegraphics[width=4.4in]{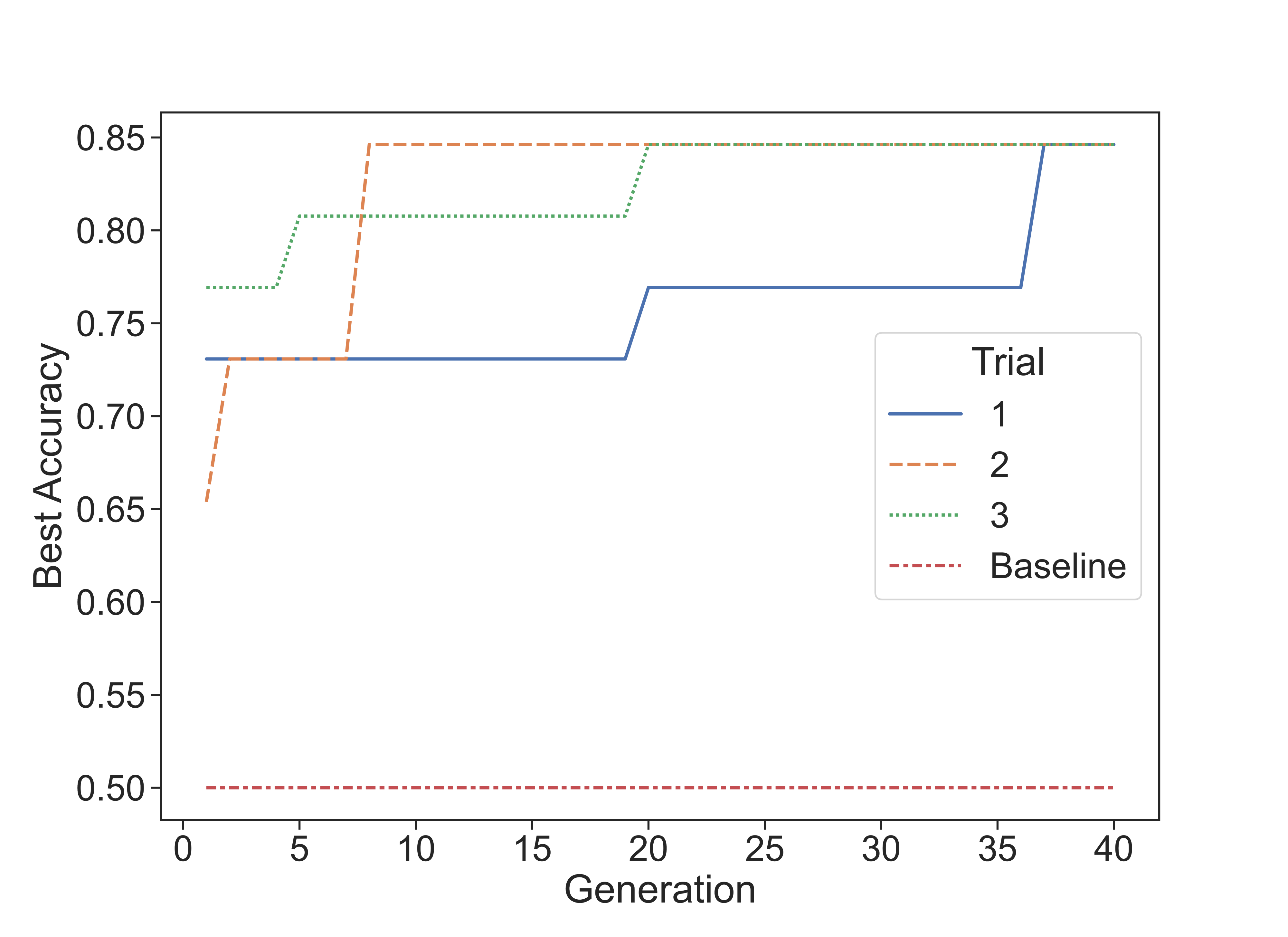}
\centering
\caption{Genetic algorithm performance for selecting best in-context examples for the GPT-3 Completion Endpoint. Results from each individual trial are compared to the baseline, which represents the expected performance of random guessing, in terms of classification accuracy on the 26-question validation set.}
\label{res:fig-genetic1}
\end{figure}

\begin{figure}[!h]
\includegraphics[width=4.4in]{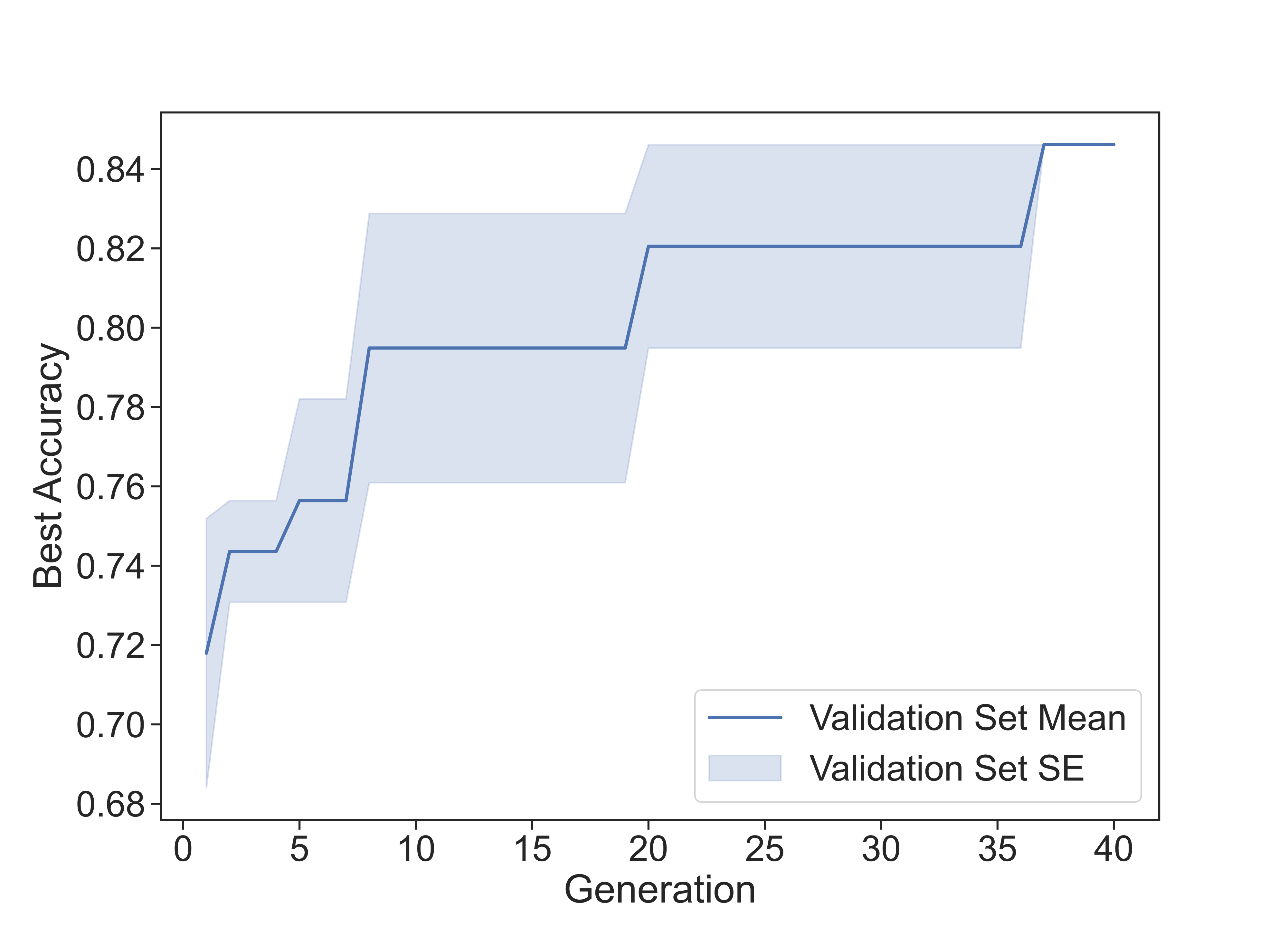}
\centering
\caption{Averaged genetic algorithm performance for selecting best in-context examples for the GPT-3 Completion Endpoint with standard error across 3 trials. Performance is measured in terms of classification accuracy on the 26-question validation set.}
\label{res:fig-genetic2}
\end{figure}

To evaluate the performance of the augmentation method for the Completion Endpoint, we first examine the changes in validation set accuracy across each generation of the genetic algorithm. Three trials of the genetic algorithm experiment were completed. Figure \ref{res:fig-genetic1} displays the validation accuracy across generations for each individual trial, while Figure \ref{res:fig-genetic2} shows the averaged results across trials, with the shaded regions representing standard error (68 percent confidence interval). 

These graphs demonstrate that, during the genetic algorithm, the \textit{validation} accuracy of the best candidate continually increased - rapidly at the start, and then more slowly as the algorithm converges. Moreover, simply selecting the best candidate from 32 random subsets, which yielded the validation accuracy at generation 0, yielded an average validation accuracy of about 72 percent. Hence, before generating any new candidates, the Completion endpoint yielded a validation accuracy comparable to the Classification Endpoint augmented with 10,000 new training examples, which obtained about 73 percent accuracy. 

Using data augmentation on the Completion Endpoint by selecting the optimal set of examples yielded a validation accuracy of about 85 percent - much better than augmentation on the Classification endpoint.  It is important to note that large spikes in accuracy often occurred in a single generation in the genetic algorithm. These spikes represented when a new best combination of training questions was found, which could have corrected several errors from the previous set at once. In addition, although generations 10 to 35 featured large standard error, by the end of each of the three trials, all simulations reached the same final best validation accuracy of about 85 percent.

However, test set results tell a different story. Table \ref{res:tab-genetic} compares the best test accuracy and validation accuracy for both the Completion Endpoint and the Classification Endpoint. Unlike the Classification Endpoint, augmenting the Completion Endpoint with a set of optimal examples failed to achieve a testing accuracy comparable to the validation accuracy. We see that the test accuracy, 67 percent, was much lower than the validation accuracy of 85 percent. 

This indicates that the in-context example set solutions produced by the genetic algorithm did \textit{not} generalize well to unseen questions. Hence, when augmented with additional in-context examples, the Completion Endpoint performed less consistently compared to the Classification Endpoint. Furthermore, since unseen questions more closely approximated those the algorithm would have been asked to classify in a practical environment, it is arguable that the augmented Classification Endpoint performed better, with a 76 percent accuracy on the test set compared to the Completion Endpoint's 67 percent accuracy.

Granted, if the algorithm were run for significantly longer than 40 generations, it is possible that a better solution could have been found. The \textit{Mean Percentage of Differing Alleles} displayed at the bottom of Table \ref{res:tab-genetic} measures the relative genetic diversity of the candidate population in generation 40. On average, 94 percent of questions in any given candidate were not shared by any other given candidate, indicating that the sets of alleles among different candidates in the final population at generation 40 contained mostly different questions. Hence, since the technique maintained genetic diversity across generations, we know it did not converge prematurely. Therefore, it could have been possible for a search to find a better solution. However, given the performance degradation between validation and test sets with the Completion Endpoint, this would probably be an inefficient use of resources. 

\begin{table}[!h]
    \caption{Comparison of augmented GPT-3 Completion Endpoint performance on data science question classification, in-context examples selected using genetic algorithm ($n_{trials}$  = 3), to augmented Classification Endpoint}
    \label{res:tab-genetic}
    \vspace{2mm}
    \centering
    \begin{tabular}{ccc}
    \toprule
        & \multicolumn{2}{c}{Endpoint} \\
        Mean Metric & Completion & Classification \\
        \midrule
        Best Validation Acc. & 0.85 & 0.73 \\
        Best Test Acc. & 0.67 ($\pm$ 0.024) & 0.76 ($\pm$ 0.017)\\
        Proportion of Differing Alleles & 0.94 ($\pm$ 0.009) & - \\
        \bottomrule
    \end{tabular}
\end{table}

\section{Discussion}
\label{discuss}

Meaning, as expressed by human language, is highly subjective. To one person, a question might be very relevant to a certain topic, while to another, it could be completely unrelated. Hence, labeling text with a category or topic depends on the complex and minute contextual associations between the words and phrases it contains. This study has demonstrated that, while GPT-3 cannot entirely overcome the limitations of this subjectivity, using data augmentation, it can capture strong enough contextual relationships between words to classify short text topics in a practical real-world deployment setting with limited data. 

This study compared data augmentation techniques for the GPT-3 Classification and Completion Endpoints, using GPT-3 to generate its own original training data examples based on observations from an existing training dataset. Augmenting the Classification Endpoint improved performance significantly, increasing validation accuracy from 49 percent to about 73 percent and increasing test set accuracy from 58 percent to 76 percent. Augmenting the Completion Endpoint yielded even better validation accuracy of about 85 percent, but this performance was inconsistent; accuracy dropped to 67 percent when evaluated on the test set, which represented unseen questions. For text classification with GPT-3, augmenting the Classification Endpoint is likely preferable to ensure consistent performance.

\subsection{Remaining Questions}

Why was there a loss in accuracy between validation and test sets for the Completion Endpoint but not for the Classification Endpoint? Firstly, the optimization performed in the Classification Endpoint only modified the hyperparameters, \textit{temperature} and \textit{max\_examples}, while for the Classification Endpoint, the genetic algorithm optimized the in-context examples themselves. If the accuracy of GPT-3 depends more on the in-context examples in the training set than on the chosen hyperparameters - which it should so that it can adapt to different problem domains - then this may have caused the genetic algorithm to overfit to the validation data. Or, the smaller training set in the Completion Endpoint may simply have been a poorer representation of the broad distribution of possible questions in general, causing worse performance. Some studies also argue that GPT-3 does not truly understand the prompt it is given \citep{Webson2021}, which supports the idea that the Completion Endpoint was not able to generalize well for classification. 

More broadly, why did data augmentation improve performance? Most likely, it worked because GPT-3 is better at generating text than classifying \citep{Brown2020}. This means that it can produce new examples that closely match the given category better than it can match a category to an example. Hence, it follows that relying more heavily on the algorithm's generative capabilities can improve its ability to identify topics correctly. In general, having more samples to describe the true population space tends to result in improved performance of machine learning models \citep{Hestness2017}. Data augmentation increases training set size; providing more samples allowed GPT-3 to learn from question-label pairs that more closely approximated the new questions that it was asked to classify. 

This improvement is also logical because it more closely approximates the thinking process used by humans. When humans make decisions, they often do not decide on a solution based on their first impulse. Rather, they brainstorm, write down different possible solutions that cover different possible scenarios and compare the best ones. They take advantage of the tools in their environment to improve reasoning, an idea which is known as ``extended cognition'' or the ``extended mind'' \citep{Clark2008}. Data augmentation acts as a sort of ``brainstorming'' process for GPT-3. By allowing GPT-3 to generate its own examples, it is allowed to more closely mimic how humans use external tools like writing to enhance the performance of their own brains. 

\subsection{Limitations and Future Work}

Of course, this study is limited by the sample size of the validation and test sets. Testing data augmentation on a practical scenario with few training examples meant the test set could only be limited to 20 questions, which obviously limits the robustness of the results. Future research on different problems with larger datasets is necessary to confirm the experimental results in this paper for GPT-3. 

Although we specifically collected and included a subset of 27 questions meant to cover a wide variety of data science topics and represent boundary cases, the set of all possible questions is broad and the impact of including these is therefore unclear. Random sampling of questions-label pairs into training and testing sets during evaluation and random selection for topic generation resulted in performance differences, as evidenced by the standard error in the results. These variations were not extremely large, though - standard errors only reached about 1-5 percent differences in accuracy. Still, additional research could address whether a starting training set of more diverse questions would also improve data augmentation efforts. 

In addition, while accuracies less than 80 percent may seem less than desirable, in reality, categories are subjective, meaning that performance will never reach very high accuracies. For example, a question that one person believed to be data science may be seen as unrelated by another person, meaning that even humans cannot achieve perfect performance. Fortunately, most of the questions that the models classified incorrectly were questions with more subjective labels. The three questions which were always misclassified by every model included the following, with their ``true'' label in square brackets:
\begin{enumerate}
    \item \textit{``Can someone help me understand how to SSH into the computing cluster on Friday?''} [Data]
    \item \textit{``Here's a link on how to make custom Jupyter notebook themes. Big Data Club-themed notebooks, anyone?''} [Data]
    \item \textit{``Are you coming to Big Data Club tomorrow?''} [Other]
\end{enumerate}

Compared to questions like ``What programming language is best for data visualization?'' these questions are less directly related to data science. In question 1, although data scientists use computing clusters, such language is not necessarily specific to data science as a field. Even though question 2 was labeled ``Data'' since Jupyter notebooks are frequently used in the data science domain, some may consider a question about modifying the appearance of the software not particularly relevant to actual data science techniques or problems. Finally, question 3 was labeled ``Other'' since attending a club meeting does not pertain to data science, but since the club in question is dedicated to data science, some might argue otherwise.

Importantly, this means that GPT-3 performs better on questions with more obvious labels and worse on questions with less obvious ones, meaning it is more likely to ensure important classifications are made correctly. That being said, for practical applications, it may be worth including data quality control measures. For example, a software application using GPT-3 for text classification could periodically poll users to both determine which questions have more subjective topics and ensure that training examples are classified correctly. Future research should study the efficacy of such techniques. %as depicted in Figure \ref{disc:bot}

% \begin{figure}
% \includegraphics[scale=0.6]{Discord-Bot.PNG}
% \centering
% \vspace*{15mm}
% \caption{Example of a data quality control measure deployed on Discord, where the training data was originally gathered. The bot polls users to ensure topics are correct and may discard training examples with no clear answer. }
% \label{disc:bot}
% \end{figure}

Finally, even though example optimization did not produce consistent results on the test set, future work may consider alternative optimization techniques for selecting the best example-label pairs to provide to GPT-3, such as reinforcement learning \citep{Arulkumaran2017}. For example, previous research has applied reinforcement learning to text generation for data augmentation \citep{Liu2020}, and similar approaches could be applied to transfer learning models like GPT-3 in examples like the one explored here. Furthermore, since the Classification Endpoint was more consistent, the optimization approach used for the Completion Endpoint might be better employed to select examples for the Classification Endpoint instead. This would yield the best of both worlds, allowing a large number of examples and ensuring only the best generated examples are included in the augmented training set. 

\section{Conclusion}
\label{conclusion}

This study finds that, through the process of data augmentation, the generative capabilities of GPT-3 allow small ($n<30$) short text data sets to be used for developing effective text classification models. Expanding the size of the training set in the GPT-3 Classification Endpoint by generating new examples using GPT-3 itself was found to increase both validation and testing accuracy. Adding 1,000 artificial examples to a Classification Endpoint model increased classification accuracy from an average baseline of 0.58 to about 0.76 - a 31 percent increase. We also explored the use of genetic algorithms for optimizing in-context examples for few-shot learning with the Completion Endpoint, but the model performed inconsistently, achieving a validation accuracy of about 85 percent but a test accuracy (on unseen questions) of only 67 percent. As such, data augmentation is more effective when using a classification-specific model.

Augmenting training data for short text classification using the generative capabilities of GPT-3 allows NLP models to be constructed even for solving problems with limited available data. This overcomes a common problem with transfer learning wherein models for specific real-world problems often require either high-quality observations or training sets larger than most practitioners would consider ``a few'' examples. If these models are coupled with methods to check, validate, or even optimize  artificial examples generated automatically, the few-shot capabilities of transfer learning models such as GPT-3 can be fully realized. Hence, using data augmentation with GPT-3 would allow businesses and organizations to more easily construct bespoke NLP models for their unique business problems.  

\section{Acknowledgments}
First, we must thank the Program in Data Science at the University of Massachusetts Dartmouth for financially supporting this project. In addition, we would also like to thank members of the University of Massachusetts Dartmouth Big Data Club for contributing questions to the data set used in this project.

%\begin{definition}{????}

%\end{definition}
\bibliographystyle{nlelike}
\bibliography{refs}

\end{document}